\documentclass{bmvc2k}
\usepackage{graphicx}
\usepackage{booktabs}
\usepackage{multirow}
\usepackage{adjustbox}
\usepackage{amsmath, amssymb, amsfonts}
\usepackage{caption}
\usepackage{tikz}
\usetikzlibrary{arrows.meta, positioning, shapes.geometric}

\usepackage[font=small,labelfont=bf]{caption}
\title{Towards Data-Efficient Medical Imaging:\\ A Generative and Semi-Supervised Framework}

\addauthor{Mosong Ma}{mosong.ma18@imperial.ac.uk}{1}
\addauthor{Tania Stathaki}{t.stathaki@imperial.ac.uk}{1}
\addauthor{Michalis Lazarou}{michalislazarou93@gmail.com}{2}

\addinstitution{
 Dept. of Electrical and Electronic Engineering\\
 Imperial College London\\
 London, UK
}
\addinstitution{
 Centre for Vision, Speech and Signal Processing\\
 University of Surrey\\
 Guildford, UK
}

\runninghead{Ma et al.}{Towards Data-Efficient Medical Imaging}

\newcommand{\eq}[1]{(\ref{eq:#1})}

\newcommand{\red}[1]{{\color{red}{#1}}}

\newcommand{\citeme}[1]{\red{[XX]}}
\newcommand{\refme}[1]{\red{(XX)}}

\newcommand{\defn}{\mathrel{:=}}

\newcommand{\vw}{\mathbf{w}}
\newcommand{\vx}{\mathbf{x}}
\newcommand{\vy}{\mathbf{y}}
\newcommand{\vz}{\mathbf{z}}

\makeatletter
\newcommand*\bdot{\mathpalette\bdot@{.7}}
\newcommand*\bdot@[2]{\mathbin{\vcenter{\hbox{\scalebox{#2}{$\m@th#1\bullet$}}}}}
\makeatother

\makeatletter
\DeclareRobustCommand\onedot{\futurelet\@let@token\@onedot}
\def\@onedot{\ifx\@let@token.\else.\null\fi\xspace}

\makeatother

\newcommand{\train}{\mathrm{train}}

\newcommand{\aug}{\mathrm{aug}}
\newcommand{\gen}{\mathrm{gen}}

\newcommand{\ours}{\textbf{SSGNet}}

\begin{document}

\maketitle

\begin{abstract}

Deep learning in medical imaging is often limited by scarce and imbalanced annotated data. We present \textbf{SSGNet}, a unified framework that combines class–specific generative modeling with iterative semi–supervised pseudo–labeling to enhance both classification and segmentation. Rather than functioning as a standalone model, SSGNet augments existing baselines by expanding training data with StyleGAN3–generated images and refining labels through iterative pseudo–labeling. Experiments across multiple medical imaging benchmarks demonstrate consistent gains in classification and segmentation performance, while Fréchet Inception Distance analysis confirms the high quality of generated samples. These results highlight SSGNet as a practical strategy to mitigate annotation bottlenecks and improve robustness in medical image analysis. The
publicly available source code can be found in \url{https://github.com/sebastianotstan/SSGNet.git}.

\end{abstract}


\section{Introduction}

Deep learning has enabled major advances in medical image analysis~\cite{Shen2017}, driving progress in classification, segmentation, and clinical decision support. However, these gains are constrained by the scarcity of annotated data and severe class imbalance~\cite{Alzubaidi2023}. Privacy regulations, annotation costs, and the rarity of certain pathologies limit dataset size and diversity, often leading to biased models with poor generalizability.

\begin{figure*}[t]
\begin{center}
    \includegraphics[width=0.9\linewidth]{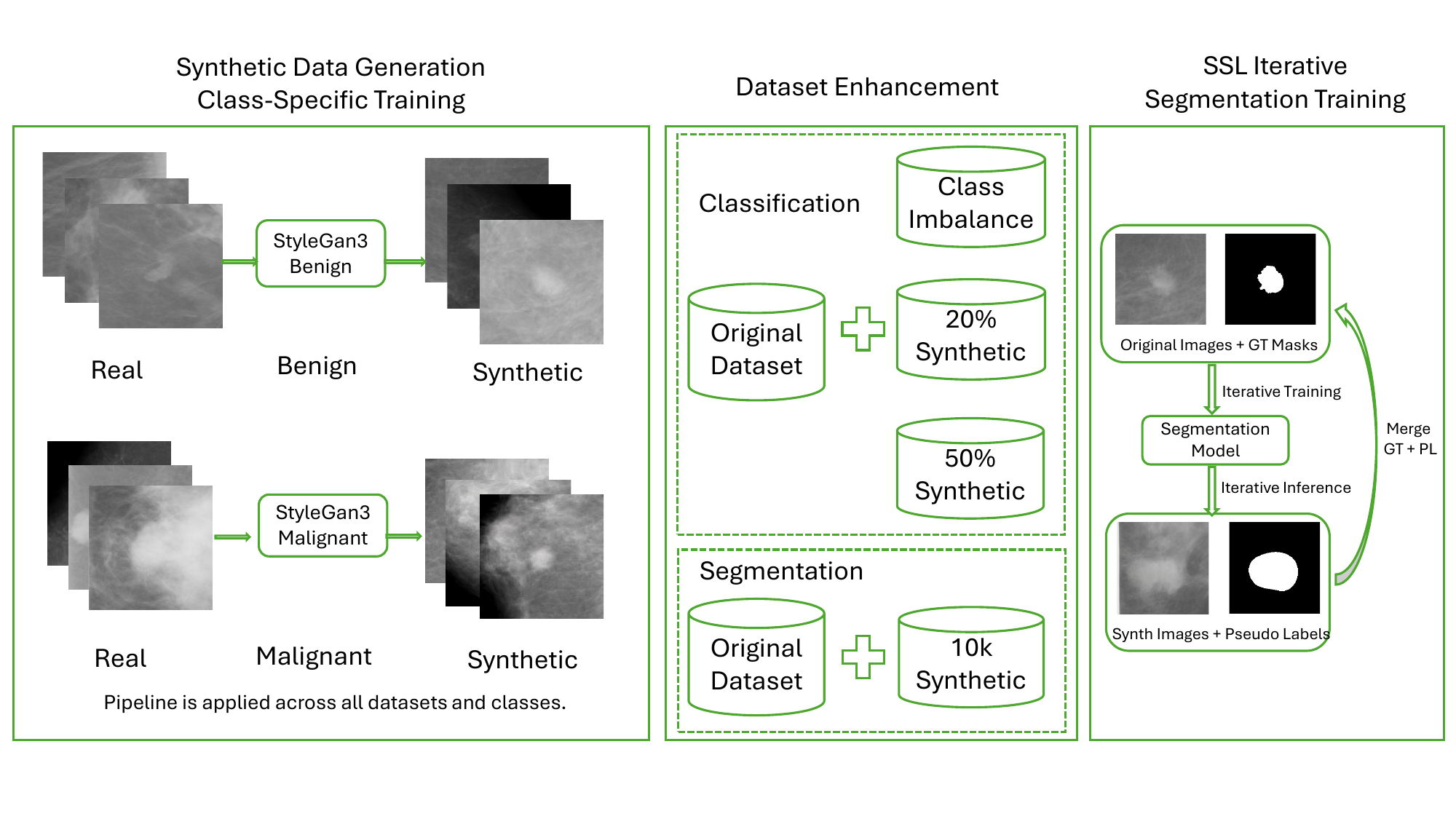}
\end{center}
    \caption{Overview of the proposed \textbf{SSGNet} pipeline. Class-specific StyleGAN3 models are trained separately for each class to generate high-quality synthetic samples, which are then used to balance or enlarge the original dataset. For classification, synthetic data are directly integrated at varying proportions. For segmentation, fixed-size 10k synthetic images are paired with pseudo-labels produced by baseline models trained on real data. These pseudo-labels are iteratively refined during training through semi-supervised learning, enabling effective use of unlabeled synthetic data.}
\label{fig:pipeline}
\end{figure*}

\subsection{Background}



Traditional augmentation techniques~\cite{Nanni2021} increase data variability but cannot fully address imbalance or scarcity. Generative models such as GANs~\cite{Goodfellow2014}, and particularly StyleGAN3~\cite{Karras2021}, can synthesize high-fidelity, diverse images, making them promising in low-data regimes. Recent work, such as PLGAN~\cite{Mao2022}, has shown that combining GAN-based augmentation with pseudo-labeling can significantly improve classification under limited supervision.

Segmentation presents an additional challenge: synthetic images typically lack corresponding ground truth masks. Semi-supervised learning strategies, especially pseudo-labeling~\cite{ssl, pseudolabel}, offer a solution by automatically generating weak labels that can be iteratively refined. This opens the door to effectively leveraging synthetic data in segmentation tasks without manual annotation.

\subsection{Contributions}
The key contributions of this work are as follows:
\begin{itemize}\setlength\itemsep{0.2em}
    \item We introduce \textbf{SSGNet}, a framework that integrates generative augmentation and semi-supervised learning to expand datasets and improve medical image analysis pipelines.
    \item We employ class-specific StyleGAN3 models to generate high-quality synthetic images, effectively expanding training data and addressing class imbalance.
    \item We implement a pseudo-labeling strategy for segmentation by applying baseline models trained on authentic data to generate masks for synthetic images, which are iteratively refined during training.
    \item We demonstrate that incorporating StyleGAN3-generated data improves both classification and segmentation performance across various medical imaging benchmarks. We further validate the quality of generated data using Fréchet Inception Distance (FID).
\end{itemize}

\section{Related Work}

\subsection{Medical Image Analysis}
Medical image analysis leverages computational methods to assist in diagnosis, treatment planning, and clinical decision-making~\cite{Ritter2011}. Despite its successes, the field faces persistent challenges due to limited access to annotated datasets, high annotation costs, and the rarity of certain conditions, resulting in class imbalance and biased models~\cite{Alzubaidi2023}. Recent research has thus explored both generative augmentation and semi-supervised learning as promising directions to alleviate these limitations.

\subsection{Datasets in Medical Image Analysis}
To benchmark methods, we employed datasets namely CBIS-DDSM~\cite{Lee2017}, Kvasir-SEG~\cite{Jha2019}, Chest X-ray~\cite{Kermany2018}, BreastMNIST~\cite{Yang2023}, ISIC 2017~\cite{Codella2017}, and ISIC 2018~\cite{Codella2019}. These datasets reflect common constraints—small training sets (often fewer than 2,000 labeled images), class imbalance, and domain-specific annotation burdens. Such limitations have motivated the integration of generative modeling and semi supervision to expand effective training sets and improve robustness.

\subsection{Generative Models for Data Augmentation}
Generative models aim to approximate data distributions and produce realistic new samples, enabling more diverse training data than traditional augmentation~\cite{Nanni2021}. GANs~\cite{Goodfellow2014} have been widely explored in medical imaging~\cite{Yi2018}, for example to synthesize chest X-rays, retinal scans, and skin lesions. StyleGAN3~\cite{Karras2021}, with its alias-free convolutions, has recently emerged as a state-of-the-art generator capable of producing high-fidelity, structurally coherent medical images. However, class imbalance during training can bias conditional GANs; several works mitigate this by training separate class-specific models. Our work follows this direction by training distinct StyleGAN3 generators for each class to ensure balanced synthetic datasets.


\subsection{Backbones for Classification and Segmentation}
Several models serve as baselines for evaluating the effectiveness of our StyleGAN3-augmented framework. 

Residual networks~\cite{He2015} remain widely used in medical classification pipelines due to their robustness and generalization. For classification, we adopt ResNet-50~\cite{He2015}, which uses residual connections to enable deeper training. 

For segmentation, we employ VM-UNet~\cite{vmunet}, which integrates Vision Mamba modules to capture long-range dependencies with linear complexity, and reference VM-UNet++~\cite{Lei2024}, an extended variant with enhanced feature aggregation though not openly available. In addition, we incorporate the Adaptive t-vMF Dice Loss framework~\cite{kato2023adaptive}, which refines the standard Dice objective by adapting similarity to better handle class imbalance and uncertain boundaries.

\subsection{Semi-Supervised Learning with Synthetic Data}
Pixel-level annotations are expensive in medical imaging, limiting the direct utility of GAN-generated images for segmentation. Semi-supervised learning (SSL)~\cite{ssl} addresses this by enabling models trained on annotated data to assign pseudo-labels to synthetic or unlabeled samples~\cite{pseudolabel}. Iterative refinement of these labels improves their reliability, allowing networks to gradually learn from weak supervision. Prior work has shown the effectiveness of such approaches in tasks like skin lesion segmentation and chest X-ray classification.

A closely related contribution is the PLGAN framework~\cite{Mao2022}, which combines GAN-based data augmentation with MixMatch-based pseudo-labeling, enhanced by contrastive learning, self-attention, and a cyclic consistency loss. PLGAN employs dual classifiers to capture both global and local features, achieving up to an 11\% accuracy improvement with limited labels.  

Our work differs in two key ways. First, because we train class-specific StyleGAN3 models, synthetic images inherently come with class labels, eliminating the need for pseudo-labeling in classification. Semi-supervised learning is therefore applied only to segmentation, where synthetic images lack masks. Second, unlike PLGAN, which focuses solely on classification, our framework jointly targets both classification and segmentation, demonstrating consistent improvements across tasks and datasets.

\section{Methods}
\subsection{Problem Definition}
We address both image classification and segmentation tasks in the context of medical imaging. We assume access to a labeled dataset $D_{\train} \defn \{(\vx_i, \vy_i)\}_{i=1}^N$, where $\vx_i$ denotes the $i^{\text{th}}$ real image and $\vy_i$ is the corresponding ground truth label. For classification, $\vy_i \in \{0, 1\}^C$ is a one-hot encoded label for $C$ classes; for segmentation, $\vy_i \in \{0,1\}^{h \times w}$ represents a binary mask where each pixel denotes the presence of the target class. $D_{\train}$ is used to train a backbone network, $f_{\theta}$, that predicts either a classification or a segmentation label, $\vy$ for an input image, $\vx$.

We then train a class-conditional StyleGAN3\cite{Karras2021} generator $g_\phi^{(k)}$ for each class $k$, using samples from $D_{\train}$ corresponding to that class. These generators are then used to synthesize class-specific images, resulting in an augmented dataset $D_{\aug} = D_{\train} \cup D_{\gen}$, where $D_{\gen}$ contains synthetic image-label pairs. For classification, the synthetic labels are known by design of the class-specific generation. For segmentation, synthetic masks are predicted by the model $f_\theta$ initially trained on $D_{\train}$.

The final dataset $D_{\aug}$ is then used to train $f_\theta$ for both tasks, leveraging real and synthetic samples in a unified manner.

\subsection{Class-Specific Synthetic Image Generation with StyleGAN3}




The StyleGAN architecture separates the generation process into two key components: a mapping network and a synthesis network. The mapping network transforms an input latent vector $\vz \sim \mathcal{Z}$, typically sampled from a standard normal distribution, into an intermediate latent code $\vw \sim \mathcal{W}$. The synthesis network then generates the final image $\hat{\vx} = g(\vz; \vw)$ by modulating convolutional kernels using $\vw$ and progressively transforming a learned constant input tensor $\vz_0 \in \mathbb{R}^{4 \times 4 \times 512}$ across multiple layers of upsampling, convolution, and nonlinearity.

Unlike earlier versions, StyleGAN3 removes explicit positional encodings and replaces upsampling operations with fully equivariant convolutions, ensuring that the network respects the continuity of the image space. The synthesis function $g$ is designed to be equivariant to smooth geometric transformations $t$, such that the transformation of the continuous input $\vz_0$ leads to a consistent transformation of the output:
\begin{equation}
g(t[\vz_0]; \vw) = t[g(\vz_0; \vw)],
\label{eq:style_invariant}
\end{equation}

This property is particularly valuable in medical imaging, where spatial coherence and fine-grained details are critical.

\subsection{Baseline Model Training}

To establish performance baselines, we first trained standard deep learning architectures for both segmentation and classification tasks using only the real, labeled training data $D_{\text{train}}$.

These baseline models serve as reference points for evaluating the benefit of synthetic data augmentation and semi-supervised learning strategies.

\paragraph{Classification.} For classification tasks, we employed a ResNet-50 backbone\cite{He2015}, a well-established convolutional neural network architecture pre-trained on ImageNet. The final classification head is replaced with a two-class output layer to match the binary nature of the datasets (e.g., benign vs malignant). Given an input image $\vx$, the model outputs class probabilities $\hat{\vy} = f_{\theta}(\vx)$, where $\hat{\vy} \in [0, 1]$. 

The network is optimized using the binary cross-entropy (BCE) loss, which penalizes discrepancies between predicted and true pixel labels:
\begin{equation}
\mathcal{L}_{\text{BCE}} = -\frac{1}{N} \sum_i \left[ \vy_i \log(\hat{\vy}_i) + (1 - \vy_i) \log(1 - \hat{\vy}_i) \right],
\label{eq:loss_bce}
\end{equation}

where $N$ is the number of pixels, $\hat{\vy}_i \in [0, 1]$ is the predicted probability for pixel $i$, $\vy_i \in \{0, 1\}$ is the ground truth label.

\paragraph{Segmentation with VM-UNet.} For the segmentation task, we adopted VM-UNet\cite{vmunet} that takes an input image $\vx$ and outputs a pixel-wise binary segmentation map $\hat{\vy} = f_{\boldsymbol{\theta}}(\vx)$, where $f_\theta$ denotes the model.

The network was trained using the BCE-Dice loss, a composite objective that combines the Binary Cross-Entropy loss and the Dice loss. While BCE penalizes pixel-wise classification errors, Dice loss measures the overlap between predicted and ground truth masks, making it particularly effective in class-imbalanced segmentation scenarios common in medical imaging.

The Dice loss is defined as in the code:
\begin{equation}
\mathcal{L}_{\text{Dice}} = 1 - \frac{1}{N} \sum_{n=1}^{N} \frac{2 \sum_{i} \hat{\mathbf{y}}_{n,i} \mathbf{y}_{n,i} + \epsilon}{\sum_{i} \hat{\mathbf{y}}_{n,i} + \sum_{i} \mathbf{y}_{n,i} + \epsilon}
\label{eq:loss_dice}
\end{equation}
where \( N \) is the batch size, \( \hat{\mathbf{y}}_{n,i} \in [0, 1] \) is the predicted probability for pixel \( i \) in sample \( n \), \( \mathbf{y}_{n,i} \in \{0, 1\} \) is the corresponding ground truth label, \( \epsilon \) is a small constant added for numerical stability.


The combined BCE-Dice loss is formulated as $\mathcal{L}_{\text{BCE-Dice}} = \mathcal{L}_{\text{BCE}} + \mathcal{L}_{\text{Dice}}$. And substituting the definitions of both components according to Equation \eq{loss_bce} and Equation \eq{loss_dice}, the full loss function becomes:
\begin{equation}
\mathcal{L}_{\text{BCE-Dice}} = -\frac{1}{N} \sum_i \left[ \vy_i \log(\hat{\vy}_i) + (1 - \vy_i) \log(1 - \hat{\vy}_i) \right] + \left(1 - \frac{1}{N} \sum_{n=1}^{N} \frac{2 \sum_{i} \hat{\mathbf{y}}_{n,i} \mathbf{y}_{n,i} + \epsilon}{\sum_{i} \hat{\mathbf{y}}_{n,i} + \sum_{i} \mathbf{y}_{n,i} + \epsilon} \right).
\label{eq:loss_bce_dice_full}
\end{equation}

This composite loss encourages the model to achieve both accurate pixel-wise predictions and coherent spatial segmentation masks.


\paragraph{Segmentation with Adaptive t-vMF Dice Loss.}
We also evaluate the Adaptive t-vMF Dice Loss~\cite{kato2023adaptive}, which extends the standard Dice formulation by incorporating the t-von Mises–Fisher (vMF) similarity as a replacement for direct overlap. The t-vMF similarity measures angular agreement between predicted probability vectors and one-hot ground truth labels, offering a more flexible similarity metric. In the adaptive variant, the sharpness parameter $\kappa$ is dynamically updated based on the class-wise Dice score at the end of each epoch, allowing the model to emphasize well-aligned classes while stabilizing learning in the presence of imbalance and ambiguous boundaries. This makes the loss more robust than BCE–Dice, particularly in medical segmentation where foreground regions are small and annotations can be noisy.

\subsection{Iterative Semi-Supervised Learning for Segmentation Tasks}

We adopted a semi-supervised learning strategy that incorporates synthetic images generated by StyleGAN3 and iteratively refines their segmentation labels. This approach allows us to augment the training data with pseudo-labeled synthetic samples, improving model generalization without requiring additional manual annotations.

Given a segmentation backbone $f_\theta$ (e.g., VM-UNet), initially trained on a labeled dataset $D_{\text{train}} = \{(\vx_i, \vy_i)\}_{i=1}^{N}$, we used the trained StyleGAN3 generator $g_\phi$ to synthesize a set of $M$ synthetic images $D_{\text{gen}} = \{\hat{\vx}_j\}_{j=1}^{M}$, where each $\hat{\vx}_j = g_\phi(\vz_j)$ and $\vz_j \sim \mathcal{Z}$ is a latent vector sampled from the input noise distribution. 

To incorporate these synthetic images into the training pipeline, we generated initial pseudo-labels using the trained segmentation model:
\begin{equation}{
\hat{\vy}_j^{(0)} = f_\theta(\hat{\vx}_j), \quad \text{for } j = 1, \ldots, M.}
\label{eq:mask_prediction}
\end{equation}
This yields the initial pseudo-labeled synthetic dataset $D_{\text{pseudo}}^{(0)} = \{(\hat{\vx}_j, \hat{\vy}_j^{(0)})\}_{j=1}^M$. We then construct an augmented training set: $D_{\text{aug}}^{(0)} = D_{\text{train}} \cup D_{\text{pseudo}}^{(0)}$
on which we retrain the segmentation model $f_\theta$. 

We adopted an iterative pseudo-label refinement strategy. After each training round $t$, we re-infer pseudo-labels on the synthetic set using the updated model:
\begin{equation}{
\hat{\vy}_j^{(t)} = f_{\theta^{(t)}}(\hat{\vx}_j)},
\label{eq:iterative_training}
\end{equation}
where $\theta^{(t)}$ are the parameters after training on $D_{\text{aug}}^{(t-1)}$. The new augmented training set becomes $D_{\text{aug}}^{(t)} = D_{\text{train}} \cup D_{\text{pseudo}}^{(t)}$

This iterative process was repeated for a fixed number of steps or until convergence. We find that even a small number of iterations (e.g., $t=2$) leads to significant improvements in segmentation performance.


\section{Experiments}

In this section, we present the experiments conducted to evaluate the performance of the proposed generative model dataset enhancement method. Specifically, we focus on the classification and segmentation tasks. We first evaluate the performance of the StyleGan3 enhanced dataset on classification tasks with the help of efficient-net and ResNet50, then move on to the segmentation experiments using VM-UNet and Adaptive t-vMF Dice Loss.

\subsection{Dataset}

We conducted experiments on six public medical image datasets: CBIS-DDSM\cite{Lee2017}, Kvasir-SEG\cite{Jha2019}, ISIC2017\cite{Codella2017}, ISIC2018\cite{Codella2019}, Chest X-ray\cite{Kermany2018} and BreastMNIST\cite{Yang2023}. Detailed dataset statistics and preprocessing steps are provided in the supplementary material.

\subsection{Hyperparameters}
To ensure stable training and high-quality image generation, we used the StyleGAN3-T\cite{Karras2021} configuration with adaptive discriminator augmentation (ADA), which mitigates overfitting in low-data regimes by applying learned augmentations to the discriminator's input. In addition, we applied dataset-specific preprocessing techniques—such as resizing, normalization, and histogram equalization—to enhance the visual quality and variability.

For synthetic augmentation in classification, we explored three strategies: balancing the classes using synthetic data, and adding synthetic samples equal to 20\% or 50\% of the balanced class size. For segmentation, we added 10,000 synthetic image–mask pairs generated using class-aware StyleGANs.

For classification training with ResNet50\cite{He2015}, we used the Adam optimizer with a decaying learning rate starting from $5 \times 10^{-3}$, a batch size of 64, and train for 100 epochs using ResNet-50. For segmentation training with VM-UNet, the model was trained for 300 epochs also using Adam with a decaying learning rate but starting at $1 \times 10^{-3}$. Early stopping was applied based on validation loss progress. While for the training of Adaptive t-vMF Dice Loss we used 200 epochs, using their innovative adaptive t-vMF dice loss and a batch size of 24 and with the TransUNet version and pretrained weights of "R50+ViT-B\_16.npz".

\subsection{Ablation Study}

We assessed the effect of synthetic data on classification and segmentation. For classification, StyleGAN3 was trained per class, and synthetic augmentation consistently improved accuracy, particularly when balancing class sizes. Additional augmentation showed further gains, varying by dataset.

For segmentation, adding 10,000 synthetic images with iterative pseudo-labeling improved performance across rounds, with diminishing returns after the second iteration.

These results confirmed the benefits of synthetic data, especially for class imbalance and limited data scenarios. Full ablation results are provided in the supplementary material.

\subsection{Experiments Results}
We evaluated the proposed SSGNet on six publicly available medical datasets: CBIS-DDSM, Kvasir-SEG, ISIC2017/2018, Chest X-ray and BreastMNIST. The experiments demonstrate the effectiveness of our method in both segmentation and classification tasks, with detailed quantitative results summarized in Table~\ref{tab:seg_comparison} and Table~\ref{tab:class_comparison}.

\paragraph{Synthesis Results}
We assessed the realism of StyleGAN3-generated images using Fréchet Inception Distance (FID)~\ref{tab:fid}. Chest X-ray models achieved the best scores (27–29), followed by dermoscopic images (35–40), and CBIS-DDSM mammograms (48–53). Polyp synthesis yielded moderately higher FIDs (53–57), though training a combined model improved to 45. Brain tumor generation remained challenging with higher scores ($>$100), likely due to limited and imbalanced training data. Conditional StyleGAN3 training was also attempted, but class imbalance reduced sample quality, confirming that class-specific training is more effective for medical data synthesis.



\vspace{0.1in}
\begin{table}[h!]
\begin{center}
\footnotesize
\begin{tabular}{|l|c|}
\hline
\textbf{Dataset / Class} & \textbf{FID} \\
\hline \hline

ChestX-ray (Normal / Pneumonia) & 27.16 / 29.31 \\
ISIC2018 / ISIC2017             & 35.40 / 40.34 \\
CBIS-DDSM (Benign / Malignant)  & 48.47 / 53.09 \\
Polyps (Normal / Polyp / Fullsize Images)  & 52.96 / 57.02 / 45.12 \\
Brain Tumor (No / Yes)          & 101.33 / 108.52 \\
\hline

\end{tabular}
\end{center}
\caption{\centering Fréchet Inception Distance (FID) scores of StyleGAN3 models trained for each dataset and class. Lower values indicate higher visual fidelity.}
\label{tab:fid}
\end{table}

\paragraph{Segmentation Results}

In segmentation tasks as shown in Table~\ref{tab:seg_comparison}, SSGNet demonstrates substantial and consistent improvements across all evaluated datasets, outperforming existing baselines such as VM-UNet, VM-UNet++, Adaptive t-vMF Dice Loss, and others in metrics including the Dice coefficient and mean Intersection over Union (mIoU). Since implementations of some variants, such as VM-UNet++, are not publicly available, we focus our experiments on two reproducible and widely used baselines—\textbf{VM-UNet} and \textbf{Adaptive t-vMF Dice Loss}. These results underscore the effectiveness of our synthetic data augmentation strategy and semi-supervised training framework. By leveraging synthetic data, we significantly enhance the model's ability to generalize in scenarios with limited labeled data.

\begin{table}[!htbp]
\begin{center}
\footnotesize
\begin{tabular}{|l|l|c|c|c|c|c|}
\hline
Dataset & Model & mIoU & F1 / Dice & Accuracy & Spe & Sen \\
\hline\hline

\multirow{7}{*}{ISIC2017} 
& UNet\cite{Ronneberger2015}            & 76.98\% & 86.99\% & 95.65\% & 97.43\% & 86.82\% \\
& UTNetV2\cite{Gao2022}                 & 77.35\% & 87.23\% & 95.84\% & 98.05\% & 84.85\% \\
& TransFuse\cite{Zhang2021}             & 79.21\% & 88.40\% & 96.17\% & 97.98\% & 87.14\% \\
& MALUNet\cite{Ruan2022}                & 78.78\% & 88.13\% & 96.18\% & \textbf{98.47\%} & 84.78\% \\
& VM-UNet++\cite{Lei2024}               & 80.49\% & 89.19\% & \textbf{96.44\%} & 98.24\% & 87.53\% \\
& VM-UNet\cite{vmunet} (reproduced)     & 78.80\% & 88.10\% & 96.10\% & 97.40\% & 88.10\% \\
& \ours + VM-UNet (Ours)                & \textbf{80.54\%} & \textbf{89.23\%} & \textbf{96.44\%} & 98.10\% & \textbf{88.15\%} \\
\cline{2-7}
& Atvmf\cite{kato2023adaptive} (reproduced) & 84.63\% & 91.43\% & 94.59\% & 96.31\% & 87.40\% \\
& \ours + Atvmf (Ours)                  & \textbf{86.90\%} & \textbf{92.81\%} & \textbf{95.45\%} & \textbf{96.76\%} & \textbf{89.98\%} \\
\hline

\multirow{9}{*}{ISIC2018}
& UNet\cite{Ronneberger2015}            & 77.86\% & 87.55\% & 94.05\% & 96.69\% & 85.86\% \\
& UNet++\cite{Zhou2018}                 & 78.31\% & 87.83\% & 94.02\% & 95.75\% & 88.65\% \\
& Att-UNet\cite{Wang2022}               & 78.43\% & 87.91\% & 94.13\% & 96.23\% & 87.60\% \\
& UTNetV2\cite{Gao2022}                 & 78.97\% & 88.25\% & 94.32\% & 96.48\% & 87.60\% \\
& SANet\cite{Wei2021}                   & 79.52\% & 88.59\% & 94.39\% & 95.97\% & 89.46\% \\
& TransFuse\cite{Zhang2021}             & 80.63\% & 89.27\% & 94.66\% & 95.74\% & 91.28\% \\
& MALUNet\cite{Ruan2022}                & 80.25\% & 89.04\% & 94.62\% & 96.19\% & 89.74\% \\
& VM-UNet++\cite{Lei2024}               & 80.17\% & 88.99\% & 94.67\% & 96.64\% & 88.52\% \\
& VM-UNet\cite{vmunet} (reproduced)     & 80.22\% & 89.04\% & 94.81\% & 96.10\% & 90.33\% \\
& \ours + VM-UNet (Ours)                & \textbf{81.54\%} & \textbf{89.83\%} & \textbf{95.02\%} & \textbf{96.70\%} & \textbf{91.80\%} \\
\cline{2-7}
& Atvmf\cite{kato2023adaptive} (reproduced) & 84.44\% & 91.42\% & 94.02\% & 96.59\% & 85.32\% \\
& \ours + Atvmf (Ours)                  & \textbf{87.10\%} & \textbf{92.96\%} & \textbf{95.11\%} & \textbf{97.45\%} & \textbf{87.20\%} \\
\hline

\multirow{4}{*}{CBIS-DDSM}
& VM-UNet\cite{vmunet} (reproduced)     & 52.51\% & 68.86\% & 94.99\% & 95.60\% & 64.22\% \\
& \ours + VM-UNet (Ours)                & \textbf{56.99\%} & \textbf{72.60\%} & \textbf{95.25\%} & \textbf{97.20\%} & \textbf{74.15\%} \\
\cline{2-7}
& Atvmf\cite{kato2023adaptive} (reproduced) & 59.63\% & 71.10\% & 87.20\% & 91.28\% & \textbf{55.33\%} \\
& \ours + Atvmf (Ours)                  & \textbf{60.98\%} & \textbf{72.22\%} & \textbf{88.51\%} & \textbf{93.10\%} & 52.60\% \\
\hline

\multirow{4}{*}{Kvasir-SEG}
& VM-UNet\cite{vmunet} (reproduced)     & 74.51\% & 85.37\% & 95.50\% & 96.22\% & 84.14\% \\
& \ours + VM-UNet (Ours)                & \textbf{76.94\%} & \textbf{87.02\%} & \textbf{95.77\%} & \textbf{97.53\%} & \textbf{87.03\%} \\
\cline{2-7}
& Atvmf\cite{kato2023adaptive} (reproduced) & 86.35\% & 92.37\% & 95.99\% & 98.70\% & 82.40\% \\
& \ours + Atvmf (Ours)                  & \textbf{86.79\%} & \textbf{92.70\%} & \textbf{96.15\%} & \textbf{98.87\%} & \textbf{82.50\%} \\
\hline

\end{tabular}
\end{center}
\caption{\centering Comparison of segmentation performance across datasets. Bold indicates the best result within each baseline model comparison (i.e., baseline vs.~baseline + SSGNet).}
\label{tab:seg_comparison}
\end{table}

\begin{figure*}[h]
\centering
\begin{minipage}[t]{0.48\linewidth}
    \centering
    \includegraphics[width=\linewidth]{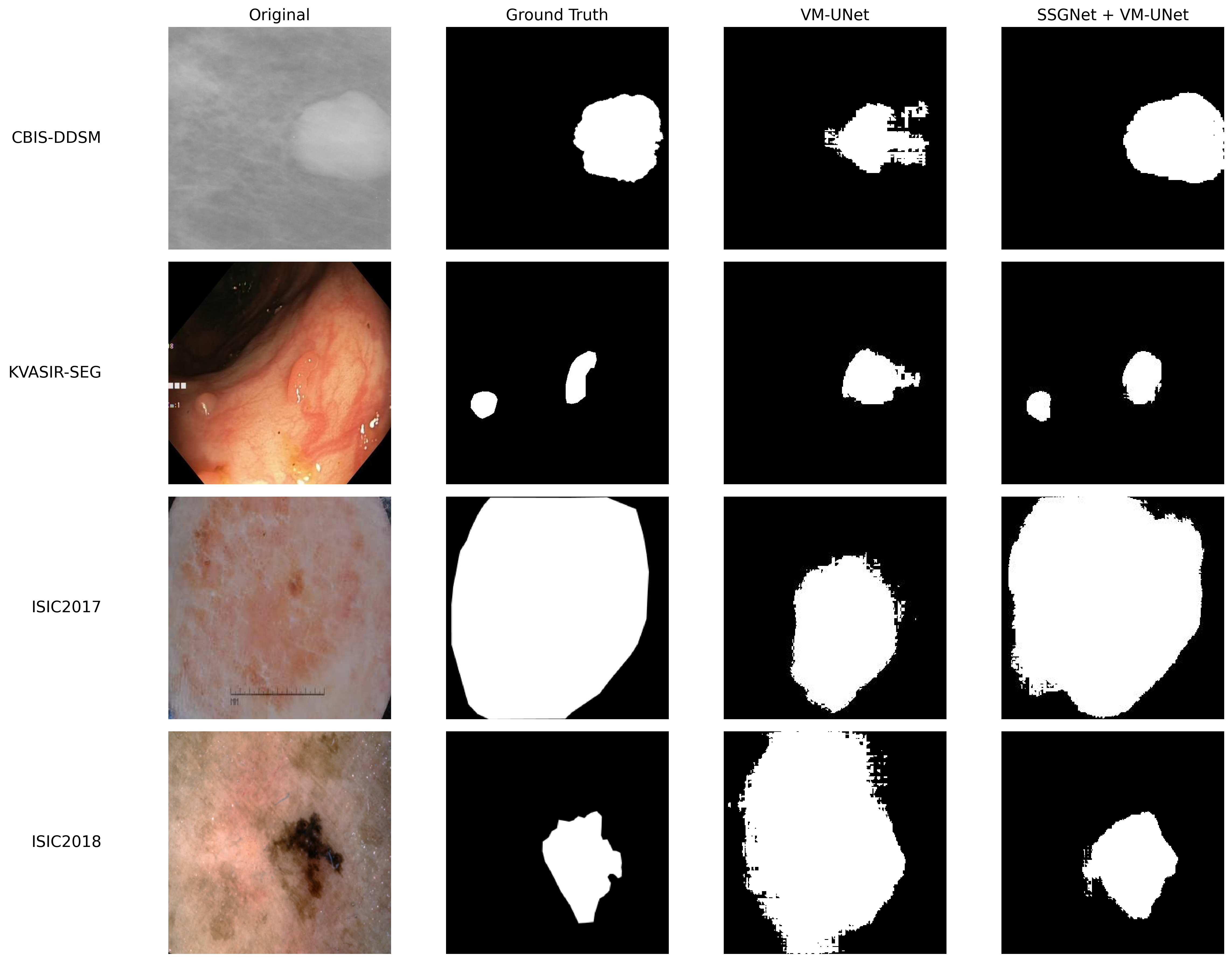}
    \caption*{(a) Comparison with VM-UNet}
\end{minipage}\hfill
\begin{minipage}[t]{0.48\linewidth}
    \centering
    \includegraphics[width=\linewidth]{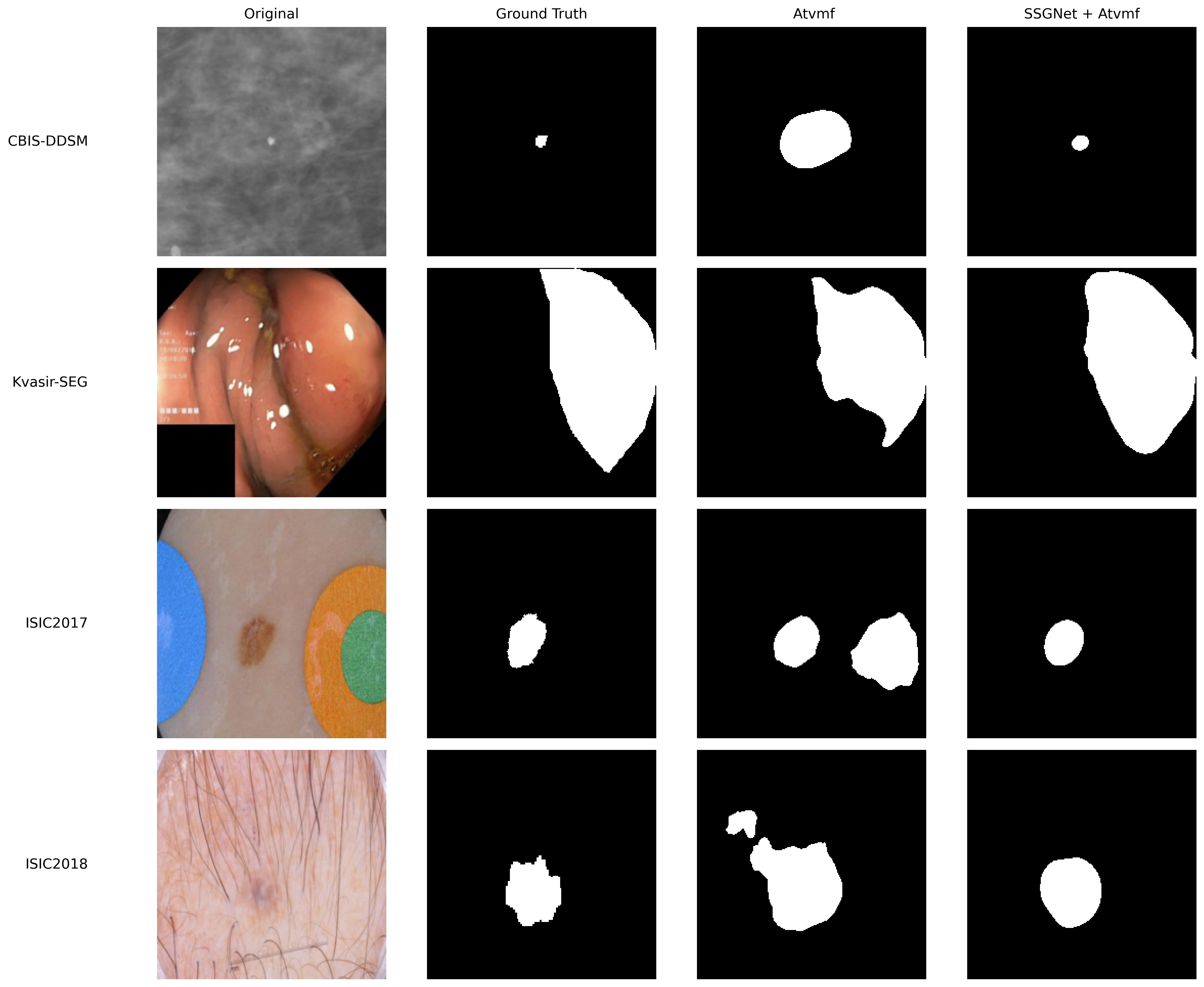}
    \caption*{(b) Comparison with Adaptive t-vMF Dice Loss}
\end{minipage}
\caption{
    Visual comparison of segmentation results across four datasets: 
    CBIS-DDSM, Kvasir-SEG, ISIC2017, and ISIC2018. 
    Left to Right in each grid: Original, Ground Truth, Baseline, and \ours +Baseline.
}
\label{fig:grid_comparison}
\end{figure*}

Notably, SSGNet consistently improves both baseline models. 
For instance, on the CBIS-DDSM dataset, it improves VM-UNet by up to \textbf{4.4\%} in mIoU and \textbf{3.6\%} in Dice, while also raising Adaptive t-vMF Dice Loss from 59.6\% to 61.0\% mIoU. 
Similar gains are observed across all other datasets, where SSGNet further boosts the already strong performance of Adaptive t-vMF Dice Loss by 0.4–1.6\% in both mIoU and Dice. 
These results demonstrate that our framework is not tied to a specific backbone or loss, but robustly enhances segmentation accuracy by leveraging generative augmentation and iterative pseudo-labeling. 
Figure~\ref{fig:grid_comparison} further illustrates qualitative improvements in segmentation quality.

\paragraph{Classification Results}
Although the main focus of this work is segmentation, we also assessed the impact of our synthetic data on classification tasks using ResNet50. As shown in Table~\ref{tab:class_comparison}, augmenting the training data with class-specific synthetic samples improves classification accuracy compared to training solely on real data. Our results show that while balancing class distributions provides a baseline improvement, further augmentation (e.g., 20\% or 50\% beyond balance) leads to additional gains, though the optimal level varies by dataset. This highlights the importance of tuning augmentation strategy per dataset.



\vspace{0.1in}
\begin{table}[h!]
\begin{center}
\footnotesize
\begin{tabular}{|l|l|c|c|}
\hline
\textbf{Dataset} & \textbf{Setting} & \textbf{Accuracy} & \textbf{Macro F1} \\
\hline \hline

Kvasir-SEG & ResNet50 & 86.0\% & 86.5\% \\
& \textbf{\ours} + ResNet50 (Ours) & \textbf{91.0\%} & \textbf{91.0\%} \\
\hline
Chest X-ray & ResNet50 & 77.0\% & 70.5\% \\
& \textbf{\ours} + ResNet50 (Ours) & \textbf{84.0\%} & \textbf{80.5\%} \\
\hline
BreastMNIST & ResNet50 & 81.0\% & 70.0\% \\
& \textbf{\ours} + ResNet50 (Ours) & \textbf{87.0\%} & \textbf{82.5\%} \\
\hline
CBIS-DDSM & ResNet50 & 60.0\% & 56.5\% \\
& \textbf{\ours} + ResNet50 (Ours) & \textbf{63.0\%} & \textbf{61.5\%} \\
\hline

\end{tabular}
\end{center}
\caption{\centering Comparison between baseline (ResNet50) and pseudo-labeled augmentation results across datasets. Bold indicates improvement over the baseline.}
\label{tab:class_comparison}
\end{table}

\section{Conclusion}
In this work, we proposed SSGNet, a semi-supervised generative framework designed to address data scarcity and class imbalance in medical imaging tasks. By leveraging synthetic data generated with class-specific StyleGAN3s and integrating it with real labeled samples using an iterative pseudo-labeling strategy, SSGNet enhances both classification and segmentation performance under limited supervision. Extensive experiments on multiple public medical datasets demonstrate that our approach consistently outperforms strong baselines trained on only real labeled data.

Our results highlight the potential of combining generative models with semi-supervised learning to improve data efficiency in medical image analysis. Future work will explore synthetic mask generation further, extend the framework to multi-class and multi-modal settings, and integrate active learning strategies to further reduce annotation costs.

\bibliography{egbib}

\begin{thebibliography}{27}
\providecommand{\natexlab}[1]{#1}
\providecommand{\url}[1]{\texttt{#1}}
\expandafter\ifx\csname urlstyle\endcsname\relax
  \providecommand{\doi}[1]{doi: #1}\else
  \providecommand{\doi}{doi: \begingroup \urlstyle{rm}\Url}\fi

\bibitem[Alzubaidi et~al.(2023)Alzubaidi, Bai, Al-Sabaawi, Santamaría,
  Albahri, Al-dabbagh, Fadhel, Manoufali, Zhang, Al-Timemy, Duan, Abdullah,
  Farhan, Lu, Gupta, Albu, Abbosh, and Gu]{Alzubaidi2023}
Laith Alzubaidi, Jinshuai Bai, Aiman Al-Sabaawi, Jose Santamaría, A.~S.
  Albahri, Bashar Sami~Nayyef Al-dabbagh, Mohammed~A. Fadhel, Mohamed
  Manoufali, Jinglan Zhang, Ali~H. Al-Timemy, Ye~Duan, Amjed Abdullah, Laith
  Farhan, Yi~Lu, Ashish Gupta, Felix Albu, Amin Abbosh, and Yuantong Gu.
\newblock A survey on deep learning tools dealing with data scarcity:
  definitions, challenges, solutions, tips, and applications.
\newblock \emph{Journal of Big Data}, 10:\penalty0 46, 4 2023.
\newblock ISSN 2196-1115.
\newblock \doi{10.1186/s40537-023-00727-2}.

\bibitem[Codella et~al.(2019)Codella, Rotemberg, Tschandl, Celebi, Dusza,
  Gutman, Helba, Kalloo, Liopyris, Marchetti, Kittler, and
  Halpern]{Codella2019}
Noel Codella, Veronica Rotemberg, Philipp Tschandl, M.~Emre Celebi, Stephen
  Dusza, David Gutman, Brian Helba, Aadi Kalloo, Konstantinos Liopyris, Michael
  Marchetti, Harald Kittler, and Allan Halpern.
\newblock Skin lesion analysis toward melanoma detection 2018: A challenge
  hosted by the international skin imaging collaboration (isic).
\newblock 2 2019.

\bibitem[Codella et~al.(2017)Codella, Gutman, Celebi, Helba, Marchetti, Dusza,
  Kalloo, Liopyris, Mishra, Kittler, and Halpern]{Codella2017}
Noel C.~F. Codella, David Gutman, M.~Emre Celebi, Brian Helba, Michael~A.
  Marchetti, Stephen~W. Dusza, Aadi Kalloo, Konstantinos Liopyris, Nabin
  Mishra, Harald Kittler, and Allan Halpern.
\newblock Skin lesion analysis toward melanoma detection: A challenge at the
  2017 international symposium on biomedical imaging (isbi), hosted by the
  international skin imaging collaboration (isic).
\newblock 10 2017.

\bibitem[Gao et~al.(2022)Gao, Zhou, Liu, Yan, Zhang, and Metaxas]{Gao2022}
Yunhe Gao, Mu~Zhou, Di~Liu, Zhennan Yan, Shaoting Zhang, and Dimitris~N.
  Metaxas.
\newblock A data-scalable transformer for medical image segmentation:
  Architecture, model efficiency, and benchmark.
\newblock 2 2022.

\bibitem[Goodfellow et~al.(2014)Goodfellow, Pouget-Abadie, Mirza, Xu,
  Warde-Farley, Ozair, Courville, and Bengio]{Goodfellow2014}
Ian~J. Goodfellow, Jean Pouget-Abadie, Mehdi Mirza, Bing Xu, David
  Warde-Farley, Sherjil Ozair, Aaron Courville, and Yoshua Bengio.
\newblock Generative adversarial networks.
\newblock 6 2014.

\bibitem[He et~al.(2015)He, Zhang, Ren, and Sun]{He2015}
Kaiming He, Xiangyu Zhang, Shaoqing Ren, and Jian Sun.
\newblock Deep residual learning for image recognition.
\newblock 12 2015.

\bibitem[Jha et~al.(2019)Jha, Smedsrud, Riegler, Halvorsen, de~Lange, Johansen,
  and Johansen]{Jha2019}
Debesh Jha, Pia~H. Smedsrud, Michael~A. Riegler, Pål Halvorsen, Thomas
  de~Lange, Dag Johansen, and Håvard~D. Johansen.
\newblock Kvasir-seg: A segmented polyp dataset.
\newblock 11 2019.

\bibitem[Karras et~al.(2021)Karras, Aittala, Laine, H\"ark\"onen, Hellsten,
  Lehtinen, and Aila]{Karras2021}
Tero Karras, Miika Aittala, Samuli Laine, Erik H\"ark\"onen, Janne Hellsten,
  Jaakko Lehtinen, and Timo Aila.
\newblock Alias-free generative adversarial networks.
\newblock In \emph{Proc. NeurIPS}, 2021.

\bibitem[Kato and Hotta(2023)]{kato2023adaptive}
Sota Kato and Kazuhiro Hotta.
\newblock Adaptive t-vmf dice loss: An effective expansion of dice loss for
  medical image segmentation.
\newblock \emph{Computers in Biology and Medicine}, page 107695, 2023.

\bibitem[Kermany et~al.(2018)Kermany, Goldbaum, Cai, Valentim, Liang, Baxter,
  McKeown, Yang, Wu, Yan, Dong, Prasadha, Pei, Ting, Zhu, Li, Hewett, Dong,
  Ziyar, Shi, Zhang, Zheng, Hou, Shi, Fu, Duan, Huu, Wen, Zhang, Zhang, Li,
  Wang, Singer, Sun, Xu, Tafreshi, Lewis, Xia, and Zhang]{Kermany2018}
Daniel~S. Kermany, Michael Goldbaum, Wenjia Cai, Carolina~C.S. Valentim,
  Huiying Liang, Sally~L. Baxter, Alex McKeown, Ge~Yang, Xiaokang Wu, Fangbing
  Yan, Justin Dong, Made~K. Prasadha, Jacqueline Pei, Magdalene~Y.L. Ting, Jie
  Zhu, Christina Li, Sierra Hewett, Jason Dong, Ian Ziyar, Alexander Shi, Runze
  Zhang, Lianghong Zheng, Rui Hou, William Shi, Xin Fu, Yaou Duan, Viet~A.N.
  Huu, Cindy Wen, Edward~D. Zhang, Charlotte~L. Zhang, Oulan Li, Xiaobo Wang,
  Michael~A. Singer, Xiaodong Sun, Jie Xu, Ali Tafreshi, M.~Anthony Lewis,
  Huimin Xia, and Kang Zhang.
\newblock Identifying medical diagnoses and treatable diseases by image-based
  deep learning.
\newblock \emph{Cell}, 172:\penalty0 1122--1131.e9, 2 2018.
\newblock ISSN 00928674.
\newblock \doi{10.1016/j.cell.2018.02.010}.

\bibitem[Lee()]{pseudolabel}
Dong-Hyun Lee.
\newblock Pseudo-label : The simple and efficient semi-supervised learning
  method for deep neural networks.
\newblock URL \url{https://www.researchgate.net/publication/280581078}.

\bibitem[Lee et~al.(2017)Lee, Gimenez, Hoogi, Miyake, Gorovoy, and
  Rubin]{Lee2017}
Rebecca~Sawyer Lee, Francisco Gimenez, Assaf Hoogi, Kanae~Kawai Miyake, Mia
  Gorovoy, and Daniel~L Rubin.
\newblock A curated mammography data set for use in computer-aided detection
  and diagnosis research.
\newblock \emph{Scientific Data}, 4:\penalty0 170177, 2017.
\newblock ISSN 2052-4463.
\newblock \doi{10.1038/sdata.2017.177}.
\newblock URL \url{https://doi.org/10.1038/sdata.2017.177}.

\bibitem[Lei and Yin(2024)]{Lei2024}
Yi~Lei and Dong Yin.
\newblock Vm-unet++: Advanced nested vision mamba unet for precise medical
  image segmentation.
\newblock pages 1012--1016. IEEE, 11 2024.
\newblock ISBN 979-8-3503-5541-3.
\newblock \doi{10.1109/ICICML63543.2024.10957912}.

\bibitem[Mao et~al.(2022)Mao, Yin, Zhang, Chen, Chang, Chen, Yu, and
  Wang]{Mao2022}
Jiawei Mao, Xuesong Yin, Guodao Zhang, Bowen Chen, Yuanqi Chang, Weibin Chen,
  Jieyue Yu, and Yigang Wang.
\newblock Pseudo-labeling generative adversarial networks for medical image
  classification.
\newblock \emph{Computers in Biology and Medicine}, 147, 2022.
\newblock ISSN 18790534.
\newblock \doi{10.1016/j.compbiomed.2022.105729}.

\bibitem[Nanni et~al.(2021)Nanni, Paci, Brahnam, and Lumini]{Nanni2021}
Loris Nanni, Michelangelo Paci, Sheryl Brahnam, and Alessandra Lumini.
\newblock Comparison of different image data augmentation approaches.
\newblock \emph{Journal of Imaging}, 7:\penalty0 254, 11 2021.
\newblock ISSN 2313-433X.
\newblock \doi{10.3390/jimaging7120254}.

\bibitem[Reddy et~al.(2018)Reddy, Viswanath, and Reddy]{ssl}
Y~C A~Padmanabha Reddy, P~Viswanath, and B~Eswara Reddy.
\newblock Semi-supervised learning: a brief review.
\newblock \emph{International Journal of Engineering \& Technology},
  7:\penalty0 81, 2 2018.
\newblock ISSN 2227-524X.
\newblock \doi{10.14419/ijet.v7i1.8.9977}.

\bibitem[Ritter et~al.(2011)Ritter, Boskamp, Homeyer, Laue, Schwier, Link, and
  Peitgen]{Ritter2011}
Felix Ritter, Tobias Boskamp, A.~Homeyer, Hendrik Laue, Michael Schwier,
  Florian Link, and H.-O. Peitgen.
\newblock Medical image analysis.
\newblock \emph{IEEE Pulse}, 2:\penalty0 60--70, 11 2011.
\newblock ISSN 2154-2287.
\newblock \doi{10.1109/MPUL.2011.942929}.

\bibitem[Ronneberger et~al.(2015)Ronneberger, Fischer, and
  Brox]{Ronneberger2015}
Olaf Ronneberger, Philipp Fischer, and Thomas Brox.
\newblock U-net: Convolutional networks for biomedical image segmentation.
\newblock 5 2015.

\bibitem[Ruan et~al.(2022)Ruan, Xiang, Xie, Liu, and Fu]{Ruan2022}
Jiacheng Ruan, Suncheng Xiang, Mingye Xie, Ting Liu, and Yuzhuo Fu.
\newblock Malunet: A multi-attention and light-weight unet for skin lesion
  segmentation.
\newblock pages 1150--1156. IEEE, 12 2022.
\newblock ISBN 978-1-6654-6819-0.
\newblock \doi{10.1109/BIBM55620.2022.9995040}.

\bibitem[Ruan et~al.(2024)Ruan, Li, and Xiang]{vmunet}
Jiacheng Ruan, Jincheng Li, and Suncheng Xiang.
\newblock Vm-unet: Vision mamba unet for medical image segmentation, 2024.
\newblock URL \url{https://arxiv.org/abs/2402.02491}.

\bibitem[Shen et~al.(2017)Shen, Wu, and Suk]{Shen2017}
Dinggang Shen, Guorong Wu, and Heung-Il Suk.
\newblock Deep learning in medical image analysis.
\newblock \emph{Annual Review of Biomedical Engineering}, 19:\penalty0
  221--248, 6 2017.
\newblock ISSN 1523-9829.
\newblock \doi{10.1146/annurev-bioeng-071516-044442}.

\bibitem[Wang et~al.(2022)Wang, Li, and Zhuang]{Wang2022}
Sihan Wang, Lei Li, and Xiahai Zhuang.
\newblock Attu-net: Attention u-net for brain tumor segmentation, 2022.

\bibitem[Wei et~al.(2021)Wei, Hu, Zhang, Li, Zhou, and Cui]{Wei2021}
Jun Wei, Yiwen Hu, Ruimao Zhang, Zhen Li, S.~Kevin Zhou, and Shuguang Cui.
\newblock Shallow attention network for polyp segmentation, 2021.

\bibitem[Yang et~al.(2023)Yang, Shi, Wei, Liu, Zhao, Ke, Pfister, and
  Ni]{Yang2023}
Jiancheng Yang, Rui Shi, Donglai Wei, Zequan Liu, Lin Zhao, Bilian Ke,
  Hanspeter Pfister, and Bingbing Ni.
\newblock Medmnist v2 - a large-scale lightweight benchmark for 2d and 3d
  biomedical image classification.
\newblock \emph{Scientific Data}, 10:\penalty0 41, 1 2023.
\newblock ISSN 2052-4463.
\newblock \doi{10.1038/s41597-022-01721-8}.

\bibitem[Yi et~al.(2018)Yi, Walia, and Babyn]{Yi2018}
Xin Yi, Ekta Walia, and Paul Babyn.
\newblock Generative adversarial network in medical imaging: A review.
\newblock 9 2018.
\newblock \doi{10.1016/j.media.2019.101552}.

\bibitem[Zhang et~al.(2021)Zhang, Liu, and Hu]{Zhang2021}
Yundong Zhang, Huiye Liu, and Qiang Hu.
\newblock Transfuse: Fusing transformers and cnns for medical image
  segmentation, 2021.

\bibitem[Zhou et~al.(2018)Zhou, Siddiquee, Tajbakhsh, and Liang]{Zhou2018}
Zongwei Zhou, Md~Mahfuzur~Rahman Siddiquee, Nima Tajbakhsh, and Jianming Liang.
\newblock Unet++: A nested u-net architecture for medical image segmentation,
  2018.

\end{thebibliography}
\newpage
\clearpage





\appendix
\begin{center}
\textbf{\Large Supplementary material}
\end{center}
\section*{Dataset Descriptions}

\textbf{Kvasir-SEG:}  
Kvasir-SEG is a publicly available dataset consisting of 1,000 gastrointestinal (GI) endoscopic images, each annotated with high-quality pixel-wise segmentation masks indicating the location of polyps. The images vary in polyp size, shape, and texture, simulating real-world diagnostic challenges. Since all images contain visible polyps, we adapted the dataset for binary classification by splitting each image into smaller non-overlapping patches and labeling them based on the presence or absence of polyps in the corresponding mask regions.

\textbf{Chest X-ray:}  
We use the Chest X-ray subset from the “Large Dataset of Labeled Optical Coherence Tomography (OCT) and Chest X-Ray Images.” This dataset contains 5,856 validated anterior-posterior chest radiographs collected from pediatric patients aged one to five years old at the Guangzhou Women and Children’s Medical Center. Each image is labeled into one of three categories: \texttt{NORMAL}, \texttt{BACTERIA}, or \texttt{VIRUS}. For simplicity and to align with common diagnostic tasks, we group \texttt{BACTERIA} and \texttt{VIRUS} together under a single class labeled \texttt{PNEUMONIA}, resulting in a binary classification setting (\texttt{NORMAL} vs. \texttt{PNEUMONIA}). The dataset is split by patient ID into distinct training and test sets to prevent data leakage. Image filenames encode the disease category, a randomized patient ID, and an intra-patient image index.

\textbf{BreastMNIST:}  
A subset of the MedMNIST v2 collection, BreastMNIST contains grayscale ultrasound images categorized for binary classification (benign vs. malignant tumors). The images are derived from real clinical settings and include subtle texture variations and shape deformations indicative of diagnostic complexity. All images are pre-centered on the region of interest to reduce irrelevant background information.

\textbf{CBIS-DDSM:}  
The Curated Breast Imaging Subset of the Digital Database for Screening Mammography (CBIS-DDSM) contains high-resolution mammography scans with annotated masks delineating tumor masses. To reduce computational overhead, we extracted only the tumor-containing regions from the large original scans. This preprocessing was applied for both classification and segmentation tasks, ensuring consistency across experiments and focusing the model on relevant lesion areas.

\textbf{ISIC2017 \& ISIC2018:}  
These datasets were released as part of the International Skin Imaging Collaboration (ISIC) challenges, aimed at advancing the segmentation and classification of skin lesions. Each image is a dermoscopic scan labeled with a binary lesion mask. ISIC2017 contains 2,000 images while ISIC2018 includes 2,594 images with greater diversity in lesion type, color, and boundary definition, offering a more challenging benchmark for model generalization.

\section*{Preprocessing}

All images were resized to a uniform resolution of $256 \times 256$ pixels to standardize input across datasets and reduce computational cost. Intensity normalization was applied channel-wise. For grayscale datasets such as BreastMNIST, CBIS-DDSM, and Chest X-ray, we duplicated the single channel across RGB to match the input requirements of pre-trained convolutional backbones.

For segmentation datasets, ground truth masks were binarized and resized to align with the resized image resolution. Minor label noise and annotation artifacts outside the primary lesion areas were suppressed using morphological operations such as erosion and dilation. In the case of CBIS-DDSM, only tumor regions were extracted and retained for both segmentation and classification to minimize memory usage and focus model learning on pathology-relevant areas.

In the Kvasir-SEG dataset, where all images contain polyps, we split each image into smaller patches and assigned binary labels depending on whether any polyp pixels were present in the corresponding patch mask. This allowed the dataset to be used for binary classification in addition to segmentation.


For classification, we evaluated three synthetic data augmentation strategies to improve performance: (1) balancing class sizes, (2) adding 20\% more samples to each class, and (3) adding 50\% more samples to each class. For segmentation, 10,000 synthetic images were added. We employed a semi-supervised training strategy involving iterative pseudo-labeling, allowing segmentation masks for synthetic images to improve progressively over multiple training rounds.

\section*{Ablation Study}

\paragraph{Classification Study}
Table~\ref{tab:classification_results} presents an ablation study assessing the impact of synthetic data augmentation on classification performance across four medical imaging datasets: Kvasir-SEG, Chest X-ray (both original and rebalanced), BreastMNIST, and CBIS-DDSM. Each dataset is evaluated under different settings: using the original training data, a balanced version of the training data (where applicable), and training with an additional 20\% or 50\% of class-conditioned synthetic samples.

Across all datasets, introducing synthetic data consistently improves classification metrics, particularly for minority or hard-to-learn classes. For instance, in the Kvasir-SEG dataset, incorporating 20\% synthetic data increased both precision and recall for negative and positive classes, leading to a 5\% absolute gain in overall accuracy (from 86.0\% to 91.0\%). Similarly, in the Chest X-ray (Original) configuration, the recall for the normal class improved markedly from 39.0\% to 59.0\% with 50\% synthetic augmentation, while the overall accuracy increased from 77.0\% to 84.0\%.

BreastMNIST also benefited from synthetic augmentation, especially for the benign class, whose recall rose from 38.0\% to 71.0\% with 50\% synthetic data, boosting the overall accuracy by 6\%. While CBIS-DDSM exhibited more modest gains, the +20\% synthetic setting still achieved a 3\% improvement in accuracy compared to the original baseline. These results highlight that our method not only enhances data balance but also improves robustness across datasets with varying levels of class imbalance and complexity.

\paragraph{Segmentation Study}

Table~\ref{tab:segmentation_results_vm} reports segmentation performance with VM-UNet on four datasets: CBIS-DDSM, Kvasir-SEG, ISIC2017, and ISIC2018, across three iterative training stages: initial training, first round of pseudo-labeling, and second round of pseudo-labeling. The metrics reported include mean Intersection over Union (mIoU), Dice coefficient (F1), and overall accuracy.

Across all datasets, we observe consistent improvement in segmentation quality with each stage of pseudo-label refinement. For instance, in CBIS-DDSM, mIoU improved from 52.51\% to 56.91\%, and Dice from 68.86\% to 72.54\%, reflecting better boundary and region accuracy. Notably, the gains are more pronounced in datasets with initially lower segmentation performance, such as CBIS-DDSM and Kvasir-SEG, suggesting that pseudo-labeling helps correct uncertainties in the initial training.

High-performing datasets like ISIC2017 and ISIC2018 also benefited from pseudo-labeling, albeit with smaller absolute gains. ISIC2018, for example, achieved a mIoU improvement of 1.32\% and a Dice improvement of 0.82\% after two rounds of refinement. These incremental gains validate the effectiveness of our iterative semi-supervised training strategy in enhancing segmentation performance, even when starting from strong baselines.

Table~\ref{tab:segmentation_results_atvmf} presents segmentation results using Adaptive t-vMF Dice Loss as the baseline with a similar setting. Reported metrics include mean Intersection over Union (mIoU) and Dice coefficient (F1).

Overall, iterative pseudo-labeling improves or stabilizes segmentation performance across most datasets. On CBIS-DDSM, mIoU increased from 71.10\% to 72.22\%, and F1 from 59.63\% to 60.98\% after two iterations, highlighting the ability of pseudo-labeling to strengthen performance in more challenging settings. For Kvasir-SEG and ISIC2017, the largest gains were observed after the first pseudo-labeling round, improving F1 by 0.44\% and 2.27\%, respectively, before slightly plateauing or decreasing in the second round. This suggests that one iteration of refinement may already capture most of the benefits for relatively high-quality baselines. 

ISIC2018 exhibited consistent improvement across both iterations, with mIoU rising from 91.42\% to 92.96\% and F1 from 84.55\% to 87.10\%. These results indicate that while the effect of iterative pseudo-labeling varies across datasets, the strategy remains generally beneficial, particularly in lower-performing datasets or when carefully limited to the first refinement stage.

\begin{table}[h!]
\centering
\footnotesize
\caption{Performance metrics across classification datasets and augmentation settings.}
\label{tab:classification_results}
\begin{adjustbox}{max width=0.95\textwidth}
\begin{tabular}{lllcccc}
\toprule
\textbf{Dataset} & \textbf{Setting} & \textbf{Class} & \textbf{Precision} & \textbf{Recall} & \textbf{F1} & \textbf{Accuracy} \\
\midrule

\multirow{6}{*}{Kvasir-SEG}
  & Original         & Negative & 95.0\% & 77.0\% & 85.0\% & \multirow{2}{*}{86.0\%} \\
  &                  & Positive & 81.0\% & 96.0\% & 88.0\% & \\
  & \textbf{+20\% Synth}      & Negative & 92.0\% & 91.0\% & 91.0\% & \multirow{2}{*}{91.0\%} \\
  &                  & Positive & 91.0\% & 92.0\% & 91.0\% & \\
  & +50\% Synth      & Negative & 90.0\% & 91.0\% & 91.0\% & \multirow{2}{*}{91.0\%} \\
  &                  & Positive & 91.0\% & 90.0\% & 91.0\% & \\

\midrule

\multirow{8}{*}{Chest X-ray (Orig)}
  & Original         & Normal    & 95.0\% & 39.0\% & 56.0\% & \multirow{2}{*}{77.0\%} \\
  &                  & Pneumonia & 74.0\% & 99.0\% & 85.0\% & \\
  & Balanced Train   & Normal    & 98.0\% & 44.0\% & 61.0\% & \multirow{2}{*}{79.0\%} \\
  &                  & Pneumonia & 75.0\% & 99.0\% & 85.0\% & \\
  & +20\% Synth      & Normal    & 93.0\% & 57.0\% & 71.0\% & \multirow{2}{*}{82.0\%} \\
  &                  & Pneumonia & 79.0\% & 97.0\% & 87.0\% & \\
  & \textbf{+50\% Synth}      & Normal    & 95.0\% & 59.0\% & 73.0\% & \multirow{2}{*}{84.0\%} \\
  &                  & Pneumonia & 80.0\% & 98.0\% & 88.0\% & \\

\midrule

\multirow{8}{*}{Chest X-ray (Rebal)}
  & Original         & Normal    & 96.0\% & 41.0\% & 58.0\% & \multirow{2}{*}{78.0\%} \\
  &                  & Pneumonia & 74.0\% & 99.0\% & 85.0\% & \\
  & Balanced Train   & Normal    & 99.0\% & 43.0\% & 60.0\% & \multirow{2}{*}{79.0\%} \\
  &                  & Pneumonia & 75.0\% & 100.0\% & 86.0\% & \\
  & \textbf{+20\% Synth}      & Normal    & 89.0\% & 67.0\% & 76.0\% & \multirow{2}{*}{85.0\%} \\
  &                  & Pneumonia & 83.0\% & 95.0\% & 89.0\% & \\
  & +50\% Synth      & Normal    & 95.0\% & 57.0\% & 71.0\% & \multirow{2}{*}{83.0\%} \\
  &                  & Pneumonia & 80.0\% & 98.0\% & 88.0\% & \\

\midrule

\multirow{8}{*}{BreastMNIST}
  & Original         & Benign     & 80.0\% & 38.0\% & 52.0\% & \multirow{2}{*}{81.0\%} \\
  &                  & Malignant  & 81.0\% & 96.0\% & 88.0\% & \\
  & Balanced Train   & Benign     & 76.0\% & 62.0\% & 68.0\% & \multirow{2}{*}{85.0\%} \\
  &                  & Malignant  & 87.0\% & 93.0\% & 90.0\% & \\
  & +20\% Synth      & Benign     & 75.0\% & 64.0\% & 69.0\% & \multirow{2}{*}{85.0\%} \\
  &                  & Malignant  & 88.0\% & 92.0\% & 90.0\% & \\
  & \textbf{+50\% Synth}      & Benign     & 77.0\% & 71.0\% & 74.0\% & \multirow{2}{*}{87.0\%} \\
  &                  & Malignant  & 90.0\% & 92.0\% & 91.0\% & \\

\midrule

\multirow{8}{*}{CBIS-DDSM}
  & Original         & Benign     & 61.0\% & 80.0\% & 69.0\% & \multirow{2}{*}{60.0\%} \\
  &                  & Malignant  & 59.0\% & 35.0\% & 44.0\% & \\
  & Balanced Train   & Benign     & 63.0\% & 70.0\% & 66.0\% & \multirow{2}{*}{60.0\%} \\
  &                  & Malignant  & 56.0\% & 48.0\% & 52.0\% & \\
  & \textbf{+20\% Synth}      & Benign     & 65.0\% & 75.0\% & 70.0\% & \multirow{2}{*}{63.0\%} \\
  &                  & Malignant  & 60.0\% & 48.0\% & 53.0\% & \\
  & +50\% Synth      & Benign     & 61.0\% & 65.0\% & 63.0\% & \multirow{2}{*}{58.0\%} \\
  &                  & Malignant  & 52.0\% & 48.0\% & 50.0\% & \\

\bottomrule
\end{tabular}
\end{adjustbox}
\end{table}

\begin{table}[htbp]
\centering
\caption{Segmentation performance using VM-UNet as baseline model across training stages: initial training, first pseudo-labeling, and second pseudo-labeling. Metrics reported are mean Intersection over Union (mIoU), Dice coefficient (F1), and accuracy.}
\label{tab:segmentation_results_vm}
\small
\begin{tabular}{lcccc}
\toprule
\textbf{Dataset} & \textbf{Training Stage} & \textbf{mIoU} & \textbf{F1 / Dice} & \textbf{Accuracy} \\
\midrule

\multirow{3}{*}{CBIS-DDSM}
  & Initial Training     & 52.51\%    & 68.86\%    & 94.97\%    \\
  & 1st Pseudo-Labeling & 56.68\%    & 72.35\%    & 95.36\%    \\
  & \textbf{2nd Pseudo-Labeling} & \textbf{56.91\%}    & \textbf{72.54\%}    & \textbf{95.44\%}    \\

\midrule
\multirow{3}{*}{Kvasir-SEG}
  & Initial Training     & 74.54\%  & 85.42\%  & 95.49\%  \\
  & 1st Pseudo-Labeling & 76.10\%  & 86.43\%  & 95.79\%  \\
  & \textbf{2nd Pseudo-Labeling} & \textbf{76.91\%}  & \textbf{86.95\%}  & \textbf{95.84\%}  \\

\midrule
\multirow{3}{*}{ISIC2017}
  & Initial Training     & 78.79\%  & 88.14\%  & 96.05\%  \\
  & 1st Pseudo-Labeling & 79.94\%  & 88.85\%  & 96.33\%  \\
  & \textbf{2nd Pseudo-Labeling} & \textbf{80.55\%}    & \textbf{89.23\%}    & \textbf{96.44\%}    \\

\midrule
\multirow{3}{*}{ISIC2018}
  & Initial Training     & 80.16\%    & 88.98\%    & 94.75\%    \\
  & 1st Pseudo-Labeling & 81.09\%    & 89.56\%    & 94.93\%    \\
  & \textbf{2nd Pseudo-Labeling} & \textbf{81.48\%}    & \textbf{89.80\%}    & \textbf{95.03\%}    \\

\bottomrule
\end{tabular}
\end{table}

\begin{table}[htbp]
\centering
\caption{Segmentation performance using Adaptive t-vMF Dice Loss as baseline model across training stages: initial training, first pseudo-labeling, and second pseudo-labeling. Metrics reported are mean Intersection over Union (mIoU) and Dice coefficient (DSC).}
\label{tab:segmentation_results_atvmf}
\small
\begin{tabular}{lcccc}
\toprule
\textbf{Dataset} & \textbf{Training Stage} & \textbf{mIoU} & \textbf{F1/Dice} & \textbf{Accuracy}\\
\midrule

\multirow{3}{*}{CBIS-DDSM}
  & Initial Training     &  71.10\% &  59.63\%  & 87.20\%\\
  & 1st Pseudo-Labeling & 71.80\%  &  60.45\%  & 87.97\%\\
      & \textbf{2nd Pseudo-Labeling} & \textbf{72.22\%}  &  \textbf{60.98\%}  & \textbf{88.51\%}\\

\midrule
\multirow{3}{*}{Kvasir-SEG}
  & Initial Training     &  92.38\% &  86.35\%  & 95.99\%\\
  & \textbf{1st Pseudo-Labeling} &  \textbf{92.70\%} &  \textbf{86.79\%} & \textbf{96.15\%} \\
  & 2nd Pseudo-Labeling & 92.53\%  &  82.44\%  & 94.61\%\\

\midrule
\multirow{3}{*}{ISIC2017}
  & Initial Training     & 91.42\%  &  84.63\%  & 94.59\%\\
  & \textbf{1st Pseudo-Labeling} & \textbf{92.81\%}  &  \textbf{86.90\%}  & \textbf{95.45\%}\\
  & 2nd Pseudo-Labeling & 91.81\%  &  85.26\%  & 94.82\%\\

\midrule
\multirow{3}{*}{ISIC2018}
  & Initial Training     & 91.42\%  &  84.55\%  & 94.02\%\\
  & 1st Pseudo-Labeling &  92.53\% & 86.37\%  & 94.82\%\\
  & \textbf{2nd Pseudo-Labeling} & \textbf{92.96\%}  &  \textbf{87.10\%}  & \textbf{95.11\%}\\

\bottomrule
\end{tabular}
\end{table}

\end{document}